\documentclass[10pt,twocolumn,letterpaper]{article}

\usepackage{iccv}
\usepackage{times}
\usepackage{epsfig}
\usepackage{graphicx}
\usepackage{amsmath}
\usepackage{amssymb}

\usepackage[pagebackref=true,breaklinks=true,letterpaper=true,colorlinks,bookmarks=false]{hyperref}

\iccvfinalcopy %

\def\iccvPaperID{xxxx} %

\ificcvfinal\fi

\begin{document}

\title{Analysis of Spatial augmentation in Self-supervised models in the purview of training and test distributions}

\author{Abhishek Jha \hspace{0,5cm}  Tinne Tuytelaars\\
ESAT-PSI, KU Leuven \\
{\tt\small firstname.lastname@esat.kuleuven.be}
}

\maketitle
\ificcvfinal\thispagestyle{empty}\fi

\begin{abstract}
In this paper, we present an empirical study of typical spatial augmentation techniques used in self-supervised representation learning methods (both contrastive and non-contrastive), namely random crop and cutout. Our contributions are: (a) we dissociate random cropping into two separate augmentations, overlap and patch, and provide a detailed analysis on the effect of area of overlap and patch size to the accuracy on down stream tasks. (b) We offer an insight into why cutout augmentation does not learn good representation, as reported in earlier literature. Finally, based on these analysis, (c) we propose a distance-based margin to the invariance loss for learning scene-centric representations for the downstream task on object-centric distribution, showing that as simple as a margin proportional to the pixel distance between the two spatial views in the scence-centric images can improve the learned representation. Our study furthers the understanding of the spatial augmentations, and the effect of the domain-gap between the training augmentations and the test distribution.

\end{abstract}
\vspace{-0.5em}

\section{Introduction}

Self-supervised learning has emerged as a powerful paradigm for representation learning, particularly in scenarios where labeled data is scarce or unavailable. By leveraging the inherent structure within unlabeled data, self-supervised methods learn to capture informative and transferable representations, which can then be utilized in various downstream tasks, such as object recognition \cite{caron2020unsupervised,richemond2020byol,chen2020exploring}, segmentation \cite{van2021unsupervised}, and scene understanding\cite{bachmann2022multimae}. Among the diverse techniques in self-supervised learning, contrastive \cite{he2019momentum, chen2020simple, misra2020self} and non-contrastive approaches \cite{richemond2020byol, chen2021exploring} have gained prominence for their ability to produce augmentation-invariant representations.

A key aspect of these methods is the use of data augmentation techniques, which are crucial for encouraging models to learn robust and generalizable features. Augmentations such as random cropping, color jittering, and cutout \cite{naji2016cementochronology} are commonly employed to create diverse views of the same image, thereby enabling the model to learn invariant features across different transformations. Despite their widespread use, the specific impact of these augmentations on the quality of the learned representations remains underexplored.

\begin{figure*}
    \centering
    {\includegraphics[width=0.9\linewidth]{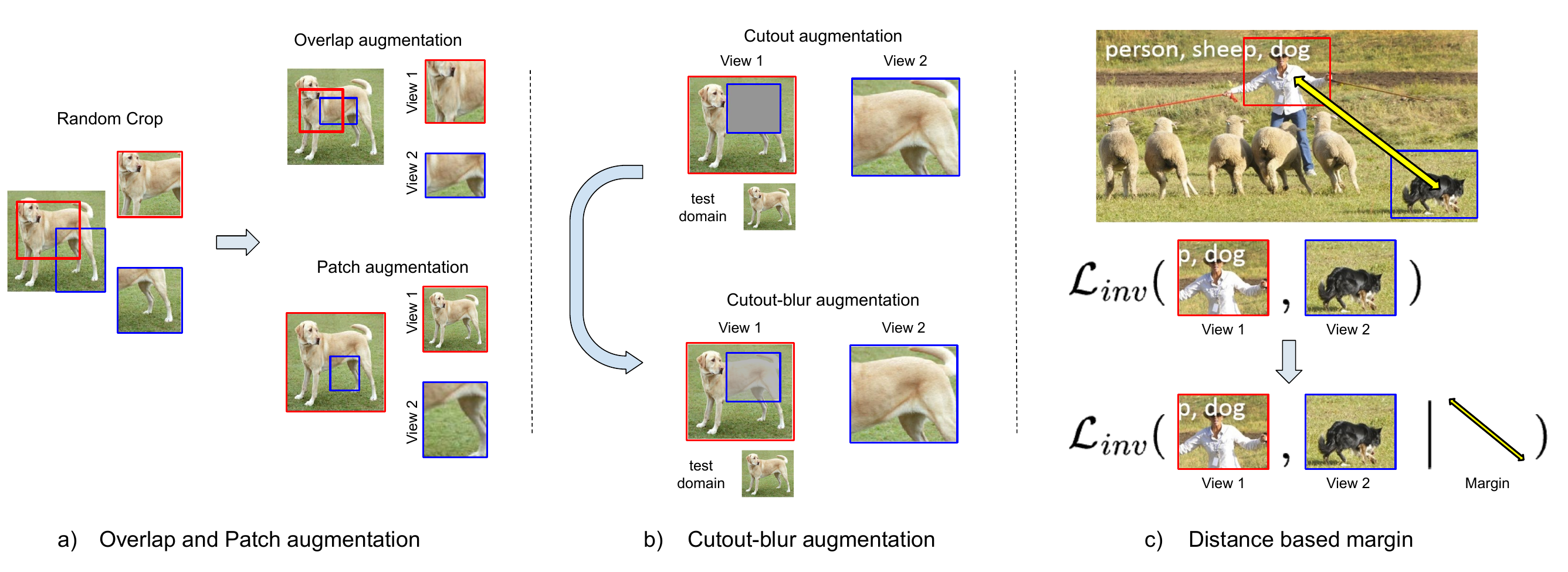}}
    \caption{\textbf{Spatial augmentation:} a) shows the overlap and patch augmentation schemes, red and blue rectangles shows the sampling regions for the two augmented views. b) shows cutout augmentation and our proposed cutout-blur augmentation. c) shows an example of scene-centric image, with multiple distinct semantic concept. Minimizing the invariance loss between views containing distinct concepts results in a noisy object-specific representation. To overcome this, we propose an invariance loss conditioned upon the distance between the two views. This distance-based margin relaxes the invariance criteria between the patches based on their inter view pixel distance.} 
    \label{fig:teaser}
    \vspace{-1em}
\end{figure*}

In particular, spatial augmentations like random cropping and cutout are of significant interest due to their ability to alter the spatial composition of images. Random cropping, which selects random parts of an image, Figure \ref{fig:teaser} (a)(left), is a staple in contrastive learning frameworks, and has been shown to be more effective than other augmentation schemes \cite{chen2020simple}. However, the effects of varying the crop size and the area of overlap between cropped regions on the learned representations have not been thoroughly investigated. Similarly, cutout as shown in Figure \ref{fig:teaser}(b)(top), which masks out random patches of the image, has been widely used in both supervised \cite{naji2016cementochronology, yun2019cutmix, zhong2020random} and self-supervised context \cite{li2021cutpaste}. However, its performance has been shown to be inferior to random crop augmentation in the context of self-supervised learning \cite{chen2020simple}.

Understanding these spatial augmentations is particularly crucial for scene-centric representation learning, where the focus is on capturing the global context of an image rather than object-specific features. While learning global context enhances performance in scene-centric tasks such as scene classification \cite{van2022discovering} and semantic segmentation \cite{bachmann2022multimae}, which require models to integrate information across different parts of the image, their performance often falls short when applied to object-centric test data \cite{van2021unsupervised}. Traditional spatial augmentation strategies, when applied to scenes containing multiple objects, uniformly sample views without respecting object boundaries, leading to a mismatch between the representations learned during training and those required for downstream object-centric tasks.

In this paper, we present an empirical study to investigate the impact of spatial augmentations on self-supervised representation learning. 
Specifically, our contributions are threefold, Figure \ref{fig:teaser}, (a) we provide a detailed analysis on the effects of overlap and patch size in random cropping augmentations on the quality of self-supervised representation; (b) we offer an explanation for the poor performance of cutout augmentation and propose a solution, cutout-blur, to fix it; and (c) we introduce a distance-based margin to the invariance loss, demonstrating its effectiveness in improving scene-centric representations. Through these contributions, we aim to enhance the understanding of spatial augmentations and how they affect the domain-gap between augmented train, and test distributions.

\section{Related work}

\textbf{Spatial Augmentation Techniques}

Random cropping and cutout are commonly used spatial augmentations. 
Random cropping forces models to learn features from different parts of an image, but its effects on representation quality as a function of crop size and overlap have not been fully explored. Studies by Chen \etal \cite{chen2020simple} and Tian \etal \cite{tian2020makes} have shown that the effectiveness of contrastive learning hinges on well-chosen augmentations. However, while the role of augmentation in general in self-supervised frameworks is well recognized, the specific effects of spatial augmentations like random cropping and cutout remain underexplored. Tian \etal \cite{tian2020makes}, provides a similar analysis as ours regarding choosing the size of random crop, however the crop size used in their work is restricted to a fixed $64\times64$ pixels, and without overlapping cases. Contrary to this, in this work we provide a more generalized analysis on variable size crops, with parameterized overlap area between them, in addition to the two other contributions. 

Cutout is one of the occlusion based augmentation scheme, where part of the image is occluded or heuristically altered, like CutMix \cite{yun2019cutmix}, CutPaste \cite{li2021cutpaste}, Random erasing\cite{zhong2020random} and so on. The pretext of our analysis comes from the empirical
study by Chen \etal \cite{chen2020simple}, which shows that cutout augmentation performs significantly worse than random cropping. In this work, we provide a systematic evaluation and possible explanation behind the inferior performance of cutout augmentation against random cropping.
Most similar to our proposed solution Cutout-blur is CutBlur \cite{yoo2020rethinking}, which swaps the cutout region of a high resolution image with that of a corresponding low resolution image and \textit{vice-versa}. Unlike CutBlur, in Cutout-blur first view is the full image with a blurred region inside it and the second view is the unblurred region corresponding to blur location in the first view. Moreover, while CutBlur addresses the problem of superresolution, in this work we provide a contemporary perspective on the underlying working mechanism of Cutout augmentation through Cutout-blur in the context of self-supervised representation learning.

Overall, while the role of augmentations in self-supervised learning is well established \cite{tian2020makes}, specific effects of spatial augmentations like random cropping and cutout require further study, particularly in scene-centric contexts. Our work addresses these gaps and proposes a novel approach to enhance scene-centric representation learning.

\section{Method and Experiments}

\begin{figure*}[t]
\centering

\begin{center}
\includegraphics[width = 0.8\textwidth]{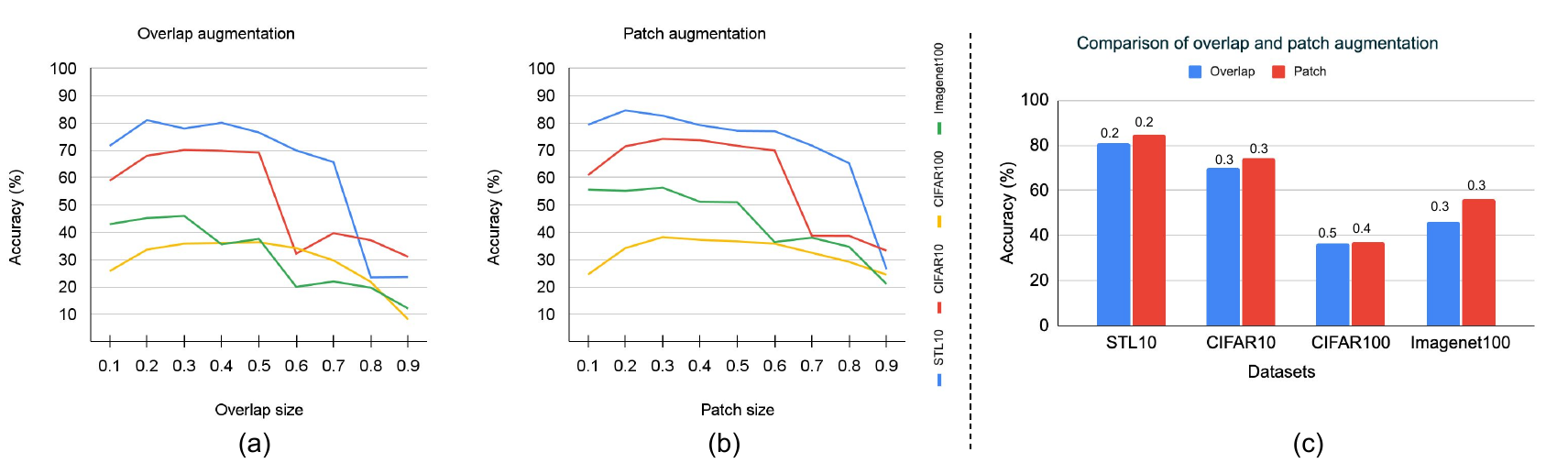}
\end{center}
\vspace{-1em}
\caption{\textbf{Analysis of Random cropping augmentation:} Evaluation of a) overlap augmentation and b) patch augmentation on STL10 \cite{coates2011analysis}, CIFAR10, CIFAR100 \cite{krizhevsky2009learning} (after 400 epochs) and Imagenet100 \cite{deng2009imagenet} (after 200 epochs). c) Comparison of best performing models from overlap to that of patch augmentation. Number on top of each bar denotes the overlap and patch sizes corresponding to the best models.}
\label{fig:spatial_augmentation_plots}
\vspace{-1em}
\end{figure*}

\subsection{Random Crop}

 Random-crop samples a crop of random size within the input image. For contrastive and non-contrastive self-supervised approaches, two random crops are generated from the same image. These crops are then independently processed through the two distinct streams. To study how the relation between the two crops affects the representation learning, we dissociate random cropping into two sub-augmentations, overlap augmentation and patch augmentation. We also investigate non-overlapping augmentations through mutually exclusive crops.

\textbf{Overlap Augmentation:}
In this augmentation technique, we sample two crops of random size from the original image such that they share an overlapping area, the size of which we define as a hyperparameter. This overlap size hyperparameter allows us to study how the learning dynamics are influenced by the shared and mutually exclusive regions. 
We define overlap size as the ratio of area of the overlapping region between the two random sized crops, and the original image. We vary this overlap size from $0.1$ to
$0.9$, and observe its effect on the downstream knn-classification performance.

We train SimSiam \cite{chen2020exploring} with ResNet50 \cite{he2015deep} as the backbone to study this new augmentation scheme, keeping other non-spatial augmentations the same as in the original SimSiam \cite{chen2020exploring}. We evaluate our models on CIFAR10/100 \cite{krizhevsky2009learning}, STL10 \cite{coates2011analysis}, and Imagenet100 \cite{deng2009imagenet} datasets.  We observe that for each of the datasets, there exists a specific overlap size that yields the best downstream performance, with an inverted-U performance curve with respect to the overlap size. This observation is similar to that of Tian \cite{tian2020makes}, however in an overlapping and more generalized crop size setting in comparison to Tian \cite{tian2020makes} where the crop sizes are fixed and mutually exclusive.
A possible explanation behind this can be that more overlap leads to more common information, and hence the network does not require learning higher-level features to minimize the invariance loss. Lower overlap sizes are noisy for the network to learn associations between the two crops. The optimal overlap is in between the two extremes, as shown in Figure \ref{fig:spatial_augmentation_plots} (a).

\textbf{Patch Augmentation:}
In patch augmentation, we keep one of the views to be the whole image, and the second view to be a patch of a fixed size, followed by other non-spatial augmentations for both the views. We control the patch size from $0.1$ to $0.9$ as the ration of patch area to that of the full image. We train our models on these two views for all four datasets, Figure \ref{fig:spatial_augmentation_plots} (b). We observe that for very small patches (size $=0.1$) and very large patches (size $\geq0.6$), the network starts converging at lower accuracies, while in the middle (size $\in [0.2, 0.4]$) we find the accuracy to be maximum. We argue that minimizing the loss for this augmentation leads to learning association between parts of the image and the whole of the image. When the patches are too small they patches capture patterns which are very generic to be associated with the object, and hence the learning signal is noisy. With very large patches, the patches capture almost the same content as the original image, and hence the representation of those two views are already in the vicinity. 
In the middle patch range, the smaller view captures the object regions that are unique to the object. The invariance loss forces these views to be closer to the full image (larger view), thereby embedding a better part-to-whole associations in the representations. We also find the trend to follows for larger dataset, like Imagenet100. 

\begin{figure}
\centering
\begin{center}
\includegraphics[width = \linewidth]{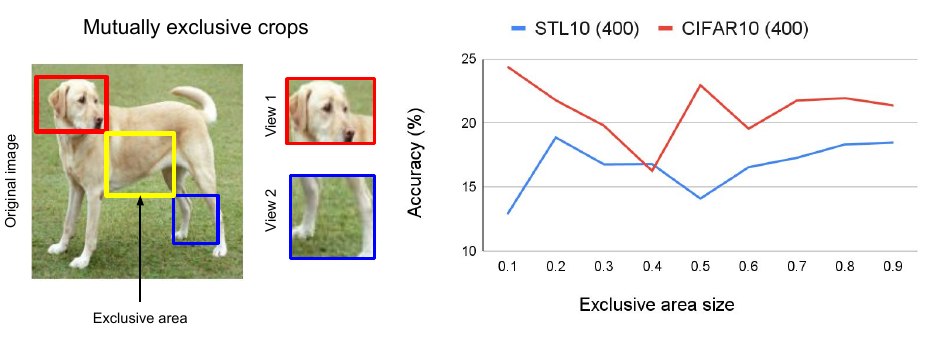}
\end{center}
\vspace{-1em}
\caption{\textbf{Mutually exclusive crops on STL10 and CIFAR10:} CIFAR10 shows a weak inverse correlation between accuracy and the area of the exclusive region, while STL10 shows no clear pattern, both measured at epoch 400.}
\label{fig:mutually_exclusive_crops}
\vspace{-2em}
\end{figure}

\textbf{Mutually exclusive crops} form the third subset of augmentations within random cropping. We parameterize this augmentation by the area of the exclusive region between the two crops of the original image (see Figure \ref{fig:mutually_exclusive_crops}). For CIFAR10, we observe a minor performance decrease as the distance between the crops increases, while no clear correlation is found for STL10.
Moreover, the overall performance of this augmentation is inferior to both overlap and patch augmentations when compared at the same epoch ($=400$), as shown in Figure \ref{fig:spatial_augmentation_plots}. This suggests that, for a randomly initialized network, long-distance associations provide a noisy learning signal.

\subsection{Cutout augmentation}
\begin{figure}
\centering
\begin{center}
\includegraphics[width=\linewidth]{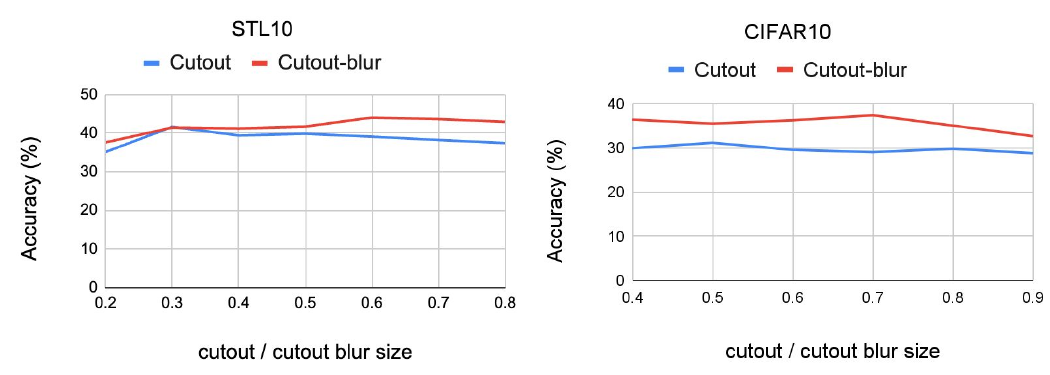}
\end{center}
\caption{\textbf{Comparison of cutout augmentation with cutout-blur augmentation:} Cutout-blur consistently outperforms cutout augmentation across different cutout sizes evaluated at the same epoch ($=100$). We do not include cutout size for which the blurring kernel is bigger than the cutout size.}
\label{fig:cutout_vs_cutout_blur}
\vspace{-1em}
\end{figure}

SimCLR \cite{chen2020simple} empirically shows that representations learned by cutout augmentation leads to an overall lower accuracy on downstream classification tasks in comparison to the random crop augmentation. We argue, this happens because of the manner cutout augmentation creates two views of the data. From our analysis on the overlap and patch augmentations, we observe that when one of the views is closer to the original distribution of the data, i.e. in the case of patch augmentation, the performance is better than when no view is in same domain as the original distribution, as shown in Figure \ref{fig:spatial_augmentation_plots} (b). In the case of cutout augmentation, none of the views are closer to the distribution of the original train and test sets, i.e. either the view contains a black patch or the other view captures only a smaller region of the whole image, resulting in an inferior performance. To test this hypothesis, we modify the  cutout augmentation by blurring the region from where the patch is cut out. This makes the first view more similar to the whole image with the noisy structural information in the cutout region. We call this augmentation cutout-blur. We compare this against the vanilla cutout augmentation, Figure \ref{fig:cutout_vs_cutout_blur}. It can be seen that by making one of the views closer to the original image, i.e. minimizing the domain-gap with the original distribution, the overall performance improves. This improvement is consistent for all cutout sizes.
Hence, the reason for lower performance of cutout augmentation can be attributed to the domain-gap between the augmented train distribution and the true distribution of the original train and test sets. It should be noted that reducing the amount of blur brings the view closer to the original image. Consequently, no blur would imply cutout augmentation, which outperforms both overlap,as discussed before, and cutout-blur, Figure \ref{fig:cutout_vs_cutout_blur}. This reinforces our hypothesis of domain-gap.

\begin{figure}
\centering

\begin{center}
\includegraphics[width = 0.8\linewidth]{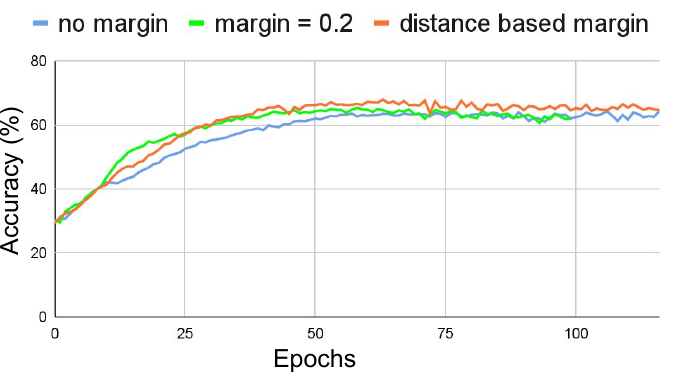}
\end{center}
\caption{Knn-evaluation on the object-centric CIFAR-10 dataset \cite{krizhevsky2009learning} comparing SimSiam trained on the scene-centric MSCOCO dataset \cite{lin2014microsoft} with no margin (vanilla), fixed-margin ($=0.2$), and distance-based margin.}
\label{fig:distance_based_margin}
\vspace{-1em}
\end{figure}

\subsection{Distance based margin}
Gansbeke \etal \cite{van2021unsupervised} show that network pretrained on scene-centric dataset perform better on scene-centric test distribution in comparison to object-centric test distribtions.
For scene-centric images, containing semantic concepts corresponding to different objects in the scene, minimizing invariance loss between different parts of the image may not be optimal for the target of learning individual concepts.  Here, we propose a distance-base margin between two views of the scene-centric data to relax the invariance minimization between potentially different semantic concepts. 
If the distance between the representations of two crops of the same image is less than the margin parametrized by the pixel distance between them, we do not backpropagate the loss gradient. 
For the two views $(x_1,x_2)$ of input $x$, with their projected and predicted representations $(p_1,z_1)$, and $(p2,z2)$ respectively, the SimSiam invariance loss can be written as:
\begin{align}
    \mathcal{L} = \sum_{i,j\in [1,2]; i\neq j}\max\big(-cos(p_i,\tt{sg}(z_j)), -1 + k\phi(x_i,x_j)\big)
\end{align}
where $i,j$ indexes the two views, $\tt{sg}$ is the stop gradient operation, $\phi$ is the $L_2$ distance in pixel space between the two crops' centers, and $k$ is a normalization constant.
We compare our method against the vanilla SimSiam trained on MSCOCO \cite{lin2014microsoft} dataset without any margin and a fixed-margin, by performing a k-nearest neighbor classification evaluation on CIFAR10 \cite{krizhevsky2009learning} dataset. Our distance-based margin outperforms the other two baselines as shown in Figure \ref{fig:distance_based_margin}, suggesting a relaxation criteria helps in improving the object-centric features given a scene-centric training set.

\section{Conclusion}
We provide an empirical analysis of two different spatial augmentation techniques, random cropping and cutout. Our experiment suggests that reducing the domain-gap between one of the augmented views of the input image with the original distribution improves the learned representations. We propose distance-based margin as an inductive bias to accommodate the domain-gap between scene-centric training and object-centric test distributions.

\textbf{Acknowledgements:} The compute used for this project has been partially supported by Vlaams Supercomputer Centrum (VSC Tier-1) and Dutch national supercomputer Snellius.

{\small
\bibliographystyle{ieee_fullname}
\bibliography{egbib}
}

\end{document}